%% file: main.tex
\documentclass[10pt,twocolumn,letterpaper]{article}

\usepackage[pagenumbers]{iccv} % To force page numbers, e.g. for an arXiv version

\input{preamble}
\input{math_commands}

\def\paperID{9621} % *** Enter the Paper ID here
\def\confName{ICCV}
\def\confYear{2025}

\title{Synchronizing Task Behavior: Aligning Multiple Tasks during Test-Time Training}

\author{
Wooseong Jeong$^{*}$, Jegyeong Cho$^{*}$, Youngho Yoon$^{*}$ and Kuk-Jin Yoon \\
Visual Intelligence Lab., KAIST, Korea\\
{\tt\small \{stk14570, j2k0618, dudgh1732, kjyoon\}@kaist.ac.kr}}

\begin{document}
\maketitle

\renewcommand{\thefootnote}{\fnsymbol{footnote}}
\footnotetext[1]{Equal contribution to this work.\\
Our source code is available at: \url{https://github.com/wooseong97/S4T-mtl}}

\input{sec/0_abstract}    
\input{sec/1_introduction}
\input{sec/2_related_works}
\input{sec/3_method}
\input{sec/4_experiments}
\input{sec/5_conclusion}

\section*{Acknowledgment}
This research was supported by the Challengeable Future Defense Technology Research and Development Program through the Agency For Defense Development(ADD) funded by the Defense Acquisition Program Administration(DAPA) in 2025(No.915102201).

{
    \small
    \bibliographystyle{ieeenat_fullname}
    \bibliography{main}
}

\input{sec/append}

\end{document}

%% file: preamble.tex
\usepackage{url} 
\usepackage{booktabs} 
\usepackage{amsfonts} 
\usepackage{nicefrac} 
\usepackage{microtype} 
\usepackage{xcolor}

\usepackage{amsmath,amsfonts,bm}
\usepackage{colortbl}
\usepackage{multirow}

\usepackage{thm-restate}

\newcommand{\ppm}{\,\scriptsize$\pm$}

\newtheorem{assumption}{Assumption}
\newtheorem{proposition}{Proposition}

\definecolor{iccvblue}{rgb}{0.21,0.49,0.74}
\usepackage[pagebackref,breaklinks,colorlinks,allcolors=iccvblue]{hyperref}

%% file: math_commands.tex
\def\eqref#1{equation~\ref{#1}}
\def\1{\bm{1}}

\DeclareMathAlphabet{\mathsfit}{\encodingdefault}{\sfdefault}{m}{sl}
\SetMathAlphabet{\mathsfit}{bold}{\encodingdefault}{\sfdefault}{bx}{n}

\newcommand{\Ls}{\mathcal{L}}

%% file: sec/0_abstract.tex
\begin{abstract}
Generalizing neural networks to unseen target domains is a significant challenge in real-world deployments. Test-time training (TTT) addresses this by using an auxiliary self-supervised task to reduce the domain gap caused by distribution shifts between the source and target. However, we find that when models are required to perform multiple tasks under domain shifts, conventional TTT methods suffer from unsynchronized task behavior, where the adaptation steps needed for optimal performance in one task may not align with the requirements of other tasks. To address this, we propose a novel TTT approach called \textbf{Synchronizing Tasks for Test-time Training (S4T)}, which enables the concurrent handling of multiple tasks. The core idea behind S4T is that predicting task relations across domain shifts is key to synchronizing tasks during test time. To validate our approach, we apply S4T to conventional multi-task benchmarks, integrating it with traditional TTT protocols. Our empirical results show that S4T outperforms state-of-the-art TTT methods across various benchmarks.
\end{abstract}

%% file: sec/1_introduction.tex
\section{Introduction}
In real-world scenarios, deep learning models are often required to perform multiple tasks simultaneously, making multi-task learning (MTL) a crucial approach for achieving both efficiency and generalization~\cite{yang2024multi, vandenhende2020mti, zhang2019pattern, huang2024going, chen2023mod, fan2022m3vit, liebel2018auxiliary, RN2, ye2022taskprompter, invpt}. Under ideal conditions, where training and test distributions remain identical, multi-task models can achieve robust performance across various tasks. However, in practical scenarios, this assumption rarely holds due to the inherent domain gap between training and test environments, which poses significant challenges for deep learning applications. Conventional domain adaptation and generalization techniques attempt to address this issue, but they are typically designed to adapt to a fixed target distribution, limiting their effectiveness in dynamic, real-world settings.

Test-time adaptation (TTA)~\cite{wang2020tent, nguyen2023tipi} and test-time training (TTT)~\cite{sun2020test, liu2021ttt++, gandelsman2022test, he2022masked, mirza2023actmad, osowiechi2023tttflow, osowiechi2024nc} have emerged as promising solutions to mitigate performance degradation due to distribution shifts. These approaches leverage target domain samples during test-time to update model parameters and enhance generalization. TTT, in particular, incorporates an auxiliary task trained on the source domain to facilitate adaptation in the target domain. By utilizing self-supervised or unsupervised learning techniques, TTT improves the model's ability to handle domain shifts during test-time.

%%%%%%%%%%%%%%%%%%%%%%%%%%%%%%%%%%%%%%%%%%%%%
\begin{figure}[t]
    \centering
    \vspace{-5pt}
    \includegraphics[width=0.99\linewidth]{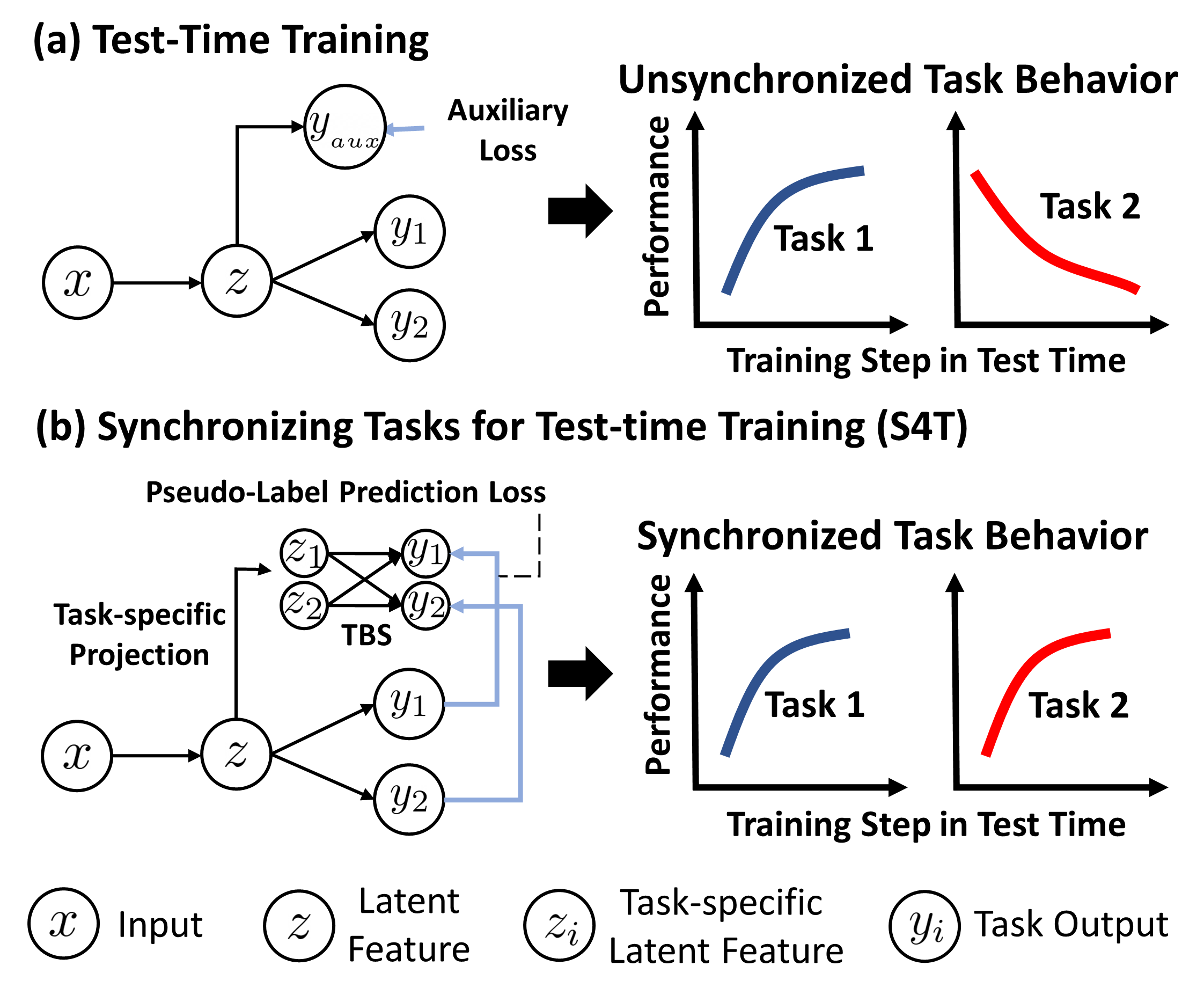}
    \vspace{-5pt}
    \caption{Comparison of conventional Test-Time Training and the proposed S4T. (a) Conventional TTT methods exhibit unsynchronized task behavior when integrated with multi-task architectures. (b) S4T leverages task relations from the source to the target domain, ensuring synchronized task behavior during test-time adaptation.}
    \label{fig:intro}
    \vspace{-8pt}
\end{figure}
%%%%%%%%%%%%%%%%%%%%%%%%%%%%%%%%%%%%%%%%%%%%%

However, as depicted in Fig.~\ref{fig:intro}-(a), conventional TTT approaches rely solely on auxiliary tasks, which presents limitations in multi-task settings. A single auxiliary task may not generalize across multiple main tasks, leading to inconsistent adaptation performance. More critically, we identify an \textbf{unsynchronization problem} in multi-task TTT, where the adaptation steps required for optimal performance in one task do not align with those needed for other tasks. This misalignment results in suboptimal adaptation across multiple tasks.

To address this challenge, we propose a novel Test-Time Training approach, called Synchronizing Tasks for Test-time Training (S4T), which leverages task relations during adaptation in the target domain rather than relying solely on independent auxiliary tasks, as shown in Fig.~\ref{fig:intro}-(b). We argue that encoding inter-task relationships in the source domain and utilizing them during test time is essential for achieving synchronized adaptation across multiple tasks. To capture these relations during training, we introduce a dedicated module separate from the main task decoders, named the \textit{Task Behavior Synchronizer (TBS)}, which utilizes task-specific latent vectors to predict task labels. Inspired by the Masked AutoEncoder (MAE)~\cite{he2022masked}, we incorporate a masking mechanism to enhance the generalizability of the learned task relations. During test-time adaptation, predictions from masked latent vectors are guided by unmasked latent representations via a pseudo-label prediction loss, effectively addressing the synchronization problem by leveraging learned task relations from the source domain. We emphasize that previous MTL and TTT approaches fail to resolve this issue, as combinations of these methods still suffer from task unsynchronization and poor multi-task performance.

To evaluate the effectiveness of our approach, we conduct experiments on several multi-task benchmarks encompassing both classification and regression tasks. Unlike prior TTT research, which primarily focuses on simple classification problems, our study explores more complex multi-task adaptation settings by incorporating diverse dense prediction tasks.
We compare S4T with existing TTT methods in various domain-shift scenarios, including NYUD-v2, Pascal-Context, and Taskonomy datasets, demonstrating that S4T consistently outperforms previous approaches. Notably, we evaluate the degree of synchronization using various metrics and reveal a positive correlation between task synchronization and multi-task performance during adaptation.

Our contributions are summarized as follows:
\begin{itemize}[leftmargin=*]
    \item We propose a novel Test-Time Training method for multi-task learning, called S4T, which explicitly addresses the unsynchronization problem by leveraging task relations as key information for synchronized test-time adaptation.
    \item Under the assumption that inter-task relations retain useful information across domain shifts, we provide a theoretical explanation of how the S4T objective reduces task loss in the target domain and maintains consistency across tasks.
    \item By comparing previous TTT methods, we demonstrate that higher synchronization enhances multi-task performance during adaptation. Additionally, S4T achieves state-of-the-art results across various domain shift scenarios.
\end{itemize}

%% file: sec/2_related_works.tex
\section{Related Work}
\textbf{Test-Time Adaptation \& Training.} Adapting deep neural networks to a target domain is challenging due to the necessity of additional burdens for collecting and labeling data in that domain.
Recent research has focused on using unlabeled data to infer the target domain's distribution, thereby narrowing the gap between source and target domains during adaptation \cite{liang2024comprehensive}. 
Test-time adaptation (TTA) and test-time training (TTT), which enable online adaptation, show broad applicability. 
A pivotal contribution in this area is TENT \cite{wang2020tent}, which uses entropy as an adaptation objective for image classification.
In computer vision, various TTA methods have been suggested to adapt off-the-shelf models during the testing phase, focusing on tasks such as image classification \cite{wang2022continual, iwasawa2021test, chen2023improved}, semantic segmentation \cite{zhang2022generalizable, volpi2022road, lee2024becotta}, and object detection \cite{fan2024test}. 
However, previous TTA methods primarily enable adaptation for classification tasks, restricting their applicability to a wider range of downstream tasks. 
On the other hand, TIPI \cite{nguyen2023tipi} enforces transformation invariance, a technique commonly used in unsupervised learning, enabling adaptation for regression tasks, although its effectiveness on these tasks has not been fully demonstrated.

TTT employs a separate self-supervised task branch \cite{sun2020test} as an auxiliary task for the main task adaptation, drawing inspiration from multi-task learning \cite{caruana1997multitask, kendall2018multi, yu2020gradient, liu2021conflict, navon2022multi, senushkin2023independent, liu2024famo, jeong2024quantifying, jeong2025selective}.
\citet{gandelsman2022test} utilizes the masked autoencoder \cite{he2022masked}, demonstrating its generalizability in handling distribution shifts during deployment.
TTT++~\cite{liu2021ttt++} preserves the statistical information of the source domain to align the test-time features through contrastive learning.
As an auxiliary self-supervised branch, TTT-MAE~\cite{gandelsman2022test} adopt Masked Autoencoder to adapt the network in the test-time domain while the normalizing flow~\cite{rezende2015variational} has been used for \citet{osowiechi2023tttflow}.
ActMad~\cite{mirza2023actmad} directly aligns the activation statistics of the test-time domain to the training domain directly, using the L1 norm.
For only the classification task, several TTT methods~\cite{su2022revisiting, hakim2023clust3, li2023robustness} adaptively update prototype clustering for each class, aligning the distribution shift.
NC-TTT~\cite{osowiechi2024nc}, the most recent TTT study, adapt model on new domain by learning to classify noisy views of projected feature maps.

\vspace{2pt}
\noindent\textbf{Task Relations in Multi-Task Learning.}
Early studies in this field employed semi-supervised learning to infer task relations~\cite{liu2007semi, zhang2009semi, wang2009semi}.
Recent works have leveraged deep neural networks across domains such as computer vision and speech recognition for more effective task relationship modeling~\cite{imran2019semi, huang2020partly, latif2020multi}. 
Several studies explicitly utilize task relations by exploiting task-specific characteristics~\cite{zamir2020robust, lu2021taskology, saha2021learning}. 
Lu et al.~\cite{lu2021taskology} regularizes depth estimation and normal vector estimation, since normal vectors can be derived from depth information. 
Saha et al.~\cite{saha2021learning} models relations between semantic segmentation and depth estimation, inspired by human perception.
Other studies address distribution shifts among tasks without explicitly analyzing task relations. 
Chen et al.~\cite{chen2020multi} applies consistency loss between similar tasks in shadow detection. 
Wang et al.~\cite{wang2022semi} uses intra-domain and inter-domain adversarial loss to align the same task across different domains. 
Li et al.~\cite{li2022learning} regularizes outputs from different paths to learn pairwise task relations. 
Nishi et al.~\cite{nishi2024joint} builds a joint-task latent space by encoding and decoding stacked task labels. 
Ye et al.~\cite{ye2024diffusionmtl} improves the diffusion model~\cite{ho2020denoising} to enhance information sharing across tasks.

The proposed S4T framework differs from previous MTL approaches in two key aspects. First, S4T leverages task relations for adaptation to domain shifts, whereas MTL methods utilize task relations primarily to improve performance within the same domain. Second, S4T introduces an additional structure specifically for adaptation, whereas prior MTL approaches integrate structural components directly into the task prediction process. This distinction enables S4T to be applied independently of existing MTL methods.

%% file: sec/3_method.tex
\begin{figure*}[t]
    \centering
    \vspace{-8pt}
    \includegraphics[width=0.99\linewidth]{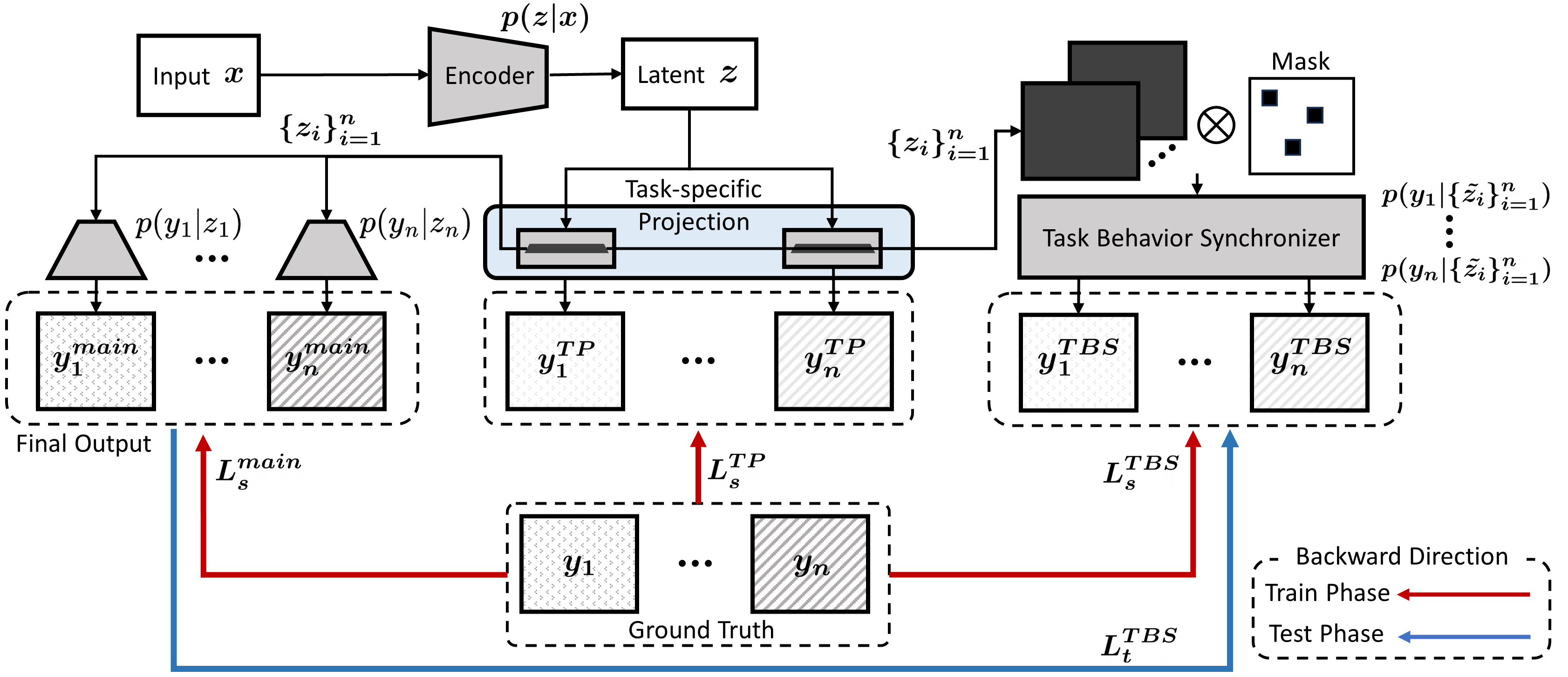}
    \vspace{-3pt}
    \caption{Overall framework.
    The input $x$ is encoded into the latent $z$ by the shared encoder $p(z|x)$. Meanwhile, task-specific projection maps $z$ to a task-specific latent $z_i$. These are stacked and fed into the task-specific decoder and TBS or passed through one more layer to predict another output $y_i^{TP}$. For each task $i$, the decoder $p(y_i|z_i)$ generates the main output $y_i^{main}$. The TBS $p(y_i|\{\Tilde{z_i}\}_{i=1}^n$) uses the masked latent $\Tilde{z_i}$ to predict the output $y_i^{TBS}$. In the train phase, each output is supervised with the ground truth $y_i$ using the loss $\Ls_s^{main}$, $\Ls_s^{TP}$ and $\Ls_s^{TBS}$, respectively. In the test phase, only $\Ls_t^{TBS}$ is minimized to adapt the framework, ensuring $y_i^{TBS}$ close to the $y_i^{main}$.  
    }
    \vspace{-2pt}
    \label{fig:main}
\end{figure*}

\section{Methods}
\subsection{Problem Definition}
Consider a domain defined by the joint distribution $p(x, y)$ with random variables $\{\mathcal{X}, \mathcal{Y}\}$, where the input data $x \sim \mathcal{X}$ and the corresponding label $y \sim \mathcal{Y}$. 
In the source domain $\{\mathcal{X}_s, \mathcal{Y}_s\}$, a deep learning network with parameters $\theta$ is trained to learn the conditional distribution $p_{\theta}(y_s|x_s)$, where $x_s \sim \mathcal{X}_s$ and $y_s \sim \mathcal{Y}_s$. 
The goal of TTT is to estimate the conditional distribution \( p(y_t | x_t) \) in the target domain \( \{\mathcal{X}_t, \mathcal{Y}_t\} \), where the distribution of \( \mathcal{X}_s \) differs from that of \( \mathcal{X}_t \), without direct access to the target domain labels \( y_t \sim \mathcal{Y}_t \).
Most of multi-task architectures \cite{riquelme2021scaling, zhang2022mixture, fan2022m3vit, mustafa2022multimodal, chen2023mod, huang2024going} use a shared encoder across tasks, generating a common latent space. We denote the random variable representing the latent space as $z_s \sim \mathcal{Z}_s$ for the source domain and $z_t \sim \mathcal{Z}_t$ for the target domain, respectively.

\subsection{Task Relations from Source to Target}
In a multi-task setting, both the source and target domains are extended to \( \{\mathcal{X}, \{\mathcal{Y}_i\}_{i=1}^{n}\} \), where \( n \) denotes the number of tasks. Similarly, the random variable representing the latent space, \( \mathcal{Z} \). As different tasks rely on different latent representations, we partition \( \mathcal{Z} \) into task-specific latent variables \( \{\mathcal{Z}_i\}_{i=1}^{n} \), where each \( \mathcal{Z}_i \) represents the conditional probability of the shared latent variable \( \mathcal{Z} \) that is relevant to task \( i \), ensuring that the overall range of the latent space satisfies \( \bigcup_{i=1}^{n} \text{Range}(\mathcal{Z}_i) \subseteq \text{Range}(\mathcal{Z}) \). We extend this notation to both the source and target domains.

Since we cannot access the ground truth of the target domain during adaptation, it is nearly impossible to predict in advance which specific information from the source domain will be useful for the target domain. As a result, most TTT approaches rely on assumptions about which information would be beneficial across domains and propose learning strategies tailored to their specific objectives.
Instead, we assume that task relations—specifically, the relationship between the set of all latent vectors and task-specific labels, denoted as \( p(y_{s,i} | z_{s,1}, z_{s,2}, \dots, z_{s,n}) \)—serve as key information that can generalize across different domains. Our first assumption is that inter-task relations remain consistent despite domain shifts, as outlined in Assumption~\ref{assumption1}.
\begin{assumption}[\textbf{Preservation of Task Relation}]
The structural dependencies between latent variables and task labels remain consistent across source and target domains:
\[
p(y_{s,i} | z_{s,1}, \dots, z_{s,n}) \approx f(p(y_{t,i} | z_{t,1}, \dots, z_{t,n}))
\]
for all \(i \in \{1, 2, \dots, n\}\), where \( f \) represents a domain-dependent transformation.
\label{assumption1}
\end{assumption}
The nature of \( f \) depends on how task relationships transform between domains. In scenarios where task relations remain structurally identical, \( f \) can be a simple distance-preserving transformation such as an isometry (e.g., rotation, translation), ensuring that pairwise distances between task representations remain unchanged. However, in cases where task relations are distorted due to domain shifts, \( f \) may be a quasi-isometry, maintaining relative distances within a bounded scale. This allows flexibility in handling domain variations where absolute distances may change, but the overall task relationships remain proportional.
While this does not cover all domain shift scenarios, we experimentally show that approaches based on this setting perform well in real-world multi-task benchmarks.

As the Masked Autoencoder (MAE) \cite{he2022masked} has shown outstanding performance in capturing a generalizable latent space of data distributions, we adapt it to approximate the probability $p(y_{s,i} | z_{s,1}, z_{s,2}, \dots, z_{s,n})$ for capturing inter-task relations. To adapt MAE, we mask the task-specific features $z_i$ with mask $\mathcal{M}_i$, where each masked task-specific feature is represented as $\tilde{z}_i \sim \tilde{\mathcal{Z}}_i$. 
These masked features are then used to jointly predict the task labels. We refer to this as Task Behavior Synchronizer (TBS). Our second assumption is that the TBS is sufficiently trained to produce final predictions with the masked task-specific latent space.
\begin{assumption}[\textbf{Sufficient Training of TBS}]
If the  Task Behavior Synchronizer is sufficiently trained, it can reliably generate task labels from the masked latent space:
\[
p(y_{i} | z_{1}, \dots, z_{n}) \approx p(y_{i} | \tilde{z}_{1}, \dots, \tilde{z}_{n})
\]
for all \(i \in \{1, 2, \dots, n\}\).
\label{assumption2}
\end{assumption}

As TBS directly receives latent vectors as input to predict task labels, we model it as a joint distribution \( p(\{z_{i}\}_{i=1}^{n}, y_j) \). 
To define the optimization objective for our TTT strategy, we quantify the discrepancy between the learnable parameters \( \theta \) of entire decoder parts of the framework and the ideal probability distribution for predicting the target task label using a distance metric \( d \), expressed as \( d(\theta, p(\{z_{t,i}\}_{i=1}^{n}, y_{t,j})) \). 
We derive the following bound:
\begin{proposition}
Under Assumptions~\ref{assumption1} and~\ref{assumption2}, the following inequality holds:
\begin{align}
    d(\theta, p(\{z_{t,i}\}_{i=1}^{n}, y_{t,j})) &\leq d(\theta, p(\{\tilde{z}_{t,i}\}_{i=1}^{n}, y_{t,j})) \label{pro:1} \\
    & \quad + \mathbb{E}_{p(\{z_{t,i}\}_{i=1}^{n}, \{\tilde{z}_{t,i}\}_{i=1}^{n})} \bigl[D\bigr]. \label{pro:2}
\end{align}
where \( D = d\left[p_{\theta}(y_{t,j}|\{z_{t,i}\}_{i=1}^{n}), p_{\theta}(y_{t,j}|\{\tilde{z}_{t,i}\}_{i=1}^{n})\right] \).
\label{proposition1}
\end{proposition}

The left-hand side of the inequality represents the loss for task $j$, which we aim to minimize. This is equivalent to the supervised learning objective on the target domain.
The first term on the right-hand side, \cref{pro:1}, represents the loss when using the masked latent vectors $\{\tilde{z}_{t,i}\}_{i=1}^{n}$. The second term, \cref{pro:2}, reflects the gap between predicting the task label with the full latent space and its masked version. 
We examine the reliability of our assumptions and the derivation of \cref{proposition1} in \cref{sup:reliability} and \cref{append:derivation}, respectively.

If task relations can be effectively learned by predicting task labels from masked task-specific features, these learned relations can facilitate adaptation across domain shifts to the target domain. This would allow us to transform $p(\{\tilde{z}_{t,i}\}_{i=1}^{n}, y_{t,j})$ back to $p(\{\tilde{z}_{s,i}\}_{i=1}^{n}, y_{s,j}))$ by Assumption~\ref{assumption1}. Therefore, our training objective is to minimize $d(\theta, p(\{\tilde{z}_{s,i}\}_{i=1}^{n}, y_{s,j})$, which supervises the predicted task labels derived from masked latent vectors using ground truth in source domain.
Our objective during test-time is given in \cref{pro:2}, and the multi-task version is as follows:
\begin{align}
    \min_{\theta} \sum_{j=1}^{n} \mathbb{E}_{p(\{z_{t,i}\}_{i=1}^{n}, \{\tilde{z}_{t,i}\}_{i=1}^{n})} [D]
    \label{eq:objective}
\end{align}
During test-time in the target domain, ground truth task labels are unavailable. Therefore, we minimize the discrepancy between task predictions made from the full set of task-specific latent vectors \( \{z_{t,i}\}_{i=1}^{n} \) and those made from the masked latent vectors \( \{\tilde{z}_{t,i}\}_{i=1}^{n} \). 
Through this adaptation process, the loss propagates to align the latent representations \( z \) in the target domain by leveraging task relations learned from the source domain.

Utilizing task relations inherently promotes synchronized task behavior during adaptation. Since the masking process forces the network to infer missing information from related tasks, it ensures that task predictions are aligned through their learned interdependencies. Specifically, minimizing the discrepancy between predictions made from masked and unmasked latent spaces encourages the network to leverage shared information across tasks. This structured dependency between task predictions helps mitigate the unsynchronization issue observed in conventional TTT approaches, where different tasks adapt at varying rates due to domain shifts.

\subsection{Test-Time Training Leveraging Task Relations}
Following the previous derivation, we implement a methodology for test-time training by capturing task relations in the source domain and leveraging them for adaptation during test time, as illustrated in Fig.~\ref{fig:main}.
The proposed framework consists of two branches: one for the main target task and another for adaptation, which includes the Task Behavior Synchronizer (TBS) to synchronize task behavior. The TBS branch is used solely for adaptation and does not contribute to the final prediction.
Since the ideal latent space for capturing task relations may differ from that for predicting outputs, we implement the separate branch to extract each task-specific latent space. 
The encoder extracts a latent vector, $z$, from the input image, $x$. 
The final output, $\{y_i^{main}\}^n_{i=1}$, are derived by passing the latent vector, $\{z_i\}^n_{i=1}$, through a decoder, and supervised with the ground truth, $\{y_i\}^n_{i=1}$.
On the other branch, the TBS captures the task relation by predicting the task labels from a set of task-specific latent vectors, $\{z_i\}^n_{i=1}$.
The set of the latent vectors are the projected versions of the latent vector $z$ pass through the task-specific projection layer.

\vspace{2pt}\noindent\textbf{Task-specific Projection.} The additional task-specific layers are used to project the latent vectors $z$ into the corresponding task-specific vectors, $\{z_i\}_{i=1}^{n}$.
The task-specific projection layers consist of two layers: one for projecting the latent vector into the task-specific vector and the subsequent layer for predicting the task labels, $y_{i}^{TP}$.
During training, this output is supervised with the ground truth to train the task-specific projection layers:
\begin{align}
    \Ls_s^{TP} = \sum_{i=1}^n \Ls_i (y_{i}^{TP}, y_{i})
\end{align}
By using task-specific projection loss, $\mathcal{L}^{TP}$, it is able to contain different task-specific information into the task-specific latent vectors.
During test time, the trained projection layers extract task-specific latent vectors in the target domain, which are used as input to TBS.

\vspace{2pt}\noindent\textbf{Task Behavior Synchronizer.} 
To reach the goal of TTT, the TBS is suggested to learn task relations. 
Similar to MAE, the TBS is implemented with ViT which predicts the both masked and unmasked regions of task label. 
The main difference is that TBS uses attention between task-specific tokens encoded from masked task-specific latent vectors, $\{\tilde{z}_i\}^n_{i=1}$.
In the training phase, the outputs from the TBS, $y_{i}^{TBS}$, are supervised with the Joint Task Prediction Loss, $\Ls^{TBS}$, using the ground truth, $y_{i}$.
The loss consists of the supervision loss for each task, $\Ls_i$:
\begin{align}
    \Ls_s^{TBS} = \sum_{i=1}^n \Ls_i (y_{i}^{TBS}, y_{i})
\end{align}
During test-time, the TBS output, $y_{i}^{TBS}$, is aligned with the main output, $y_{i}^{main}$, which lowers the upper bound of the objective~\cref{eq:objective} on the target domain.
Therefore, the total framework is trained with the Pseudo-label Prediction Loss $\Ls_t^{TBS}$ and this loss function is summarized as follows:
\begin{align}
    \Ls_t^{TBS} = \sum_{i=1}^n \Ls_i (y_{i}^{TBS}, y_{i}^{main})
\end{align}

In summary, during the train phase, the overall framework is trained by minimizing the following total loss (red arrows in the Fig.~\ref{fig:main}):
\begin{align}
    \Ls_s^{Total} 
    &= \Ls_s^{main}  + \lambda^{TBS}\Ls_s^{TBS} + \lambda^{TP}\Ls_s^{TP} 
\end{align}
Each $\lambda^{TBS}$ and $\lambda^{TP}$ denotes the loss weight for $\Ls_s^{TBS}$ and $\Ls_s^{TP}$, respectively.
In the test-time phase, following the typical TTT approaches, the framework is only trained with the Pseudo-label Prediction Loss (blue arrows in the Fig.~\ref{fig:main}):
\begin{align}
    \Ls_t^{Total} 
    = \Ls_t^{TBS}
    = \sum_{i=1}^n \Ls_i (y_{i}^{TBS}, y_{i}^{main})
\end{align}

%% file: sec/4_experiments.tex
\begin{table*}[t!]
    \centering
    \renewcommand{\arraystretch}{1.30}
    \caption{Comparison of multi-task performance from Taskonomy to NYUD-v2 and PASCAL-Context across different tasks for S4T, against previous TTA and TTT methods.}
    \vspace{-5pt}
    \resizebox{\linewidth}{!}
    {
    \begin{tabular}{l|cccc|>{\columncolor[HTML]{C9D6F7}}c|ccc|>{\columncolor[HTML]{C9D6F7}}c} 
    \hline \rowcolor{gray!30}
        Dataset & \multicolumn{5}{c|}{Taskonomy $\rightarrow$ NYUD-v2} & \multicolumn{4}{c}{Taskonomy $\rightarrow$ PASCAL-Context} \\ \hline \rowcolor{gray!30}
        Task & \multicolumn{1}{c}{Semseg} & \multicolumn{1}{c}{Depth} & \multicolumn{1}{c}{Normal} & \multicolumn{1}{c}{Edge} & \multicolumn{1}{c|}{$\triangle_{TTT}$} & \multicolumn{1}{c}{Semseg} & \multicolumn{1}{c}{Normal} & \multicolumn{1}{c}{Edge} & \multicolumn{1}{c}{$\triangle_{TTT}$} \\ \rowcolor{gray!30}
        Metric & mIoU $\uparrow$ & RMSE $\downarrow$ & mErr $\downarrow$ & RMSE $\downarrow$ & $\%$ $\uparrow$ & mIoU $\uparrow$ & mErr $\downarrow$ & RMSE $\downarrow$ & $\%$ $\uparrow$ \\ \hline
        Base & 29.31\ppm6.30e-2 & 1.179\ppm8.00e-3 & 61.32\ppm8.20e-1 & 0.1443\ppm7.10e-5 & +0.00 & 27.08\ppm1.40e-2 & 63.46\ppm9.50e-1 & 0.1185\ppm7.10e-5 & 0.00 \\ \hline
        TENT~\cite{wang2020tent} & 40.42\ppm1.09e+0 & 1.056\ppm1.70e-2 & 56.09\ppm3.21e+0 & 0.1441\ppm2.10e-4 & +14.26 & 40.65\ppm1.30e-1 & 58.76\ppm2.05e+0 & 0.1183\ppm9.00e-5 & +19.26 \\  
        TIPI~\cite{nguyen2023tipi} & 48.12\ppm1.78e+0 & 1.029\ppm6.50e-2 & 55.71\ppm4.93e-1 & 0.1440\ppm6.40e-5 & +21.57 & 43.01\ppm1.30e-5 & 39.03\ppm2.30e-4 & 0.1186\ppm1.40e-8 & +32.41 \\  \hline
        TTT~\cite{sun2020test} & 41.31\ppm4.46e-1 & 1.061\ppm1.00e-3 & 47.54\ppm4.31e-1 & 0.1440\ppm2.83e-5 & +18.43 & 39.29\ppm2.30e-1 & 33.76\ppm9.04e-1 & 0.1183\ppm4.00e-5 & +30.70 \\  
        TTT++~\cite{liu2021ttt++} & 43.97\ppm1.11e+0 & 1.107\ppm1.50e-2 & 46.71\ppm6.40e-2 & 0.1440\ppm2.60e-5 & +20.05 & 37.26\ppm5.00e-2 & 36.87\ppm4.50e-2 & 0.1183\ppm2.40e-4 & +26.55 \\  
        TTTFlow~\cite{osowiechi2023tttflow} & 52.75\ppm7.50e-2 & 1.075\ppm1.00e-3 & 46.02\ppm1.25e-1 & 0.1442\ppm4.70e-5 & +28.47 & 38.73\ppm2.45e-1 & 43.30\ppm7.55e-1 & 0.1184\ppm4.70e-5 & +27.97 \\ 
        ClusT3~\cite{hakim2023clust3} & 41.88\ppm2.06e+0 & 1.102\ppm1.40e-2 & 46.52\ppm7.00e-3 & 0.1440\ppm7.30e-6 & +18.44 & 33.26\ppm1.20e-5 & 35.31\ppm3.70e-5 & 0.1184\ppm3.60e-8 & +22.43 \\ 
        ActMAD~\cite{mirza2023actmad} & 27.62\ppm1.61e-1 & 1.193\ppm8.00e-3 & 56.53\ppm2.31e+0 & 0.1444\ppm3.00e-5 & +0.20 & 22.09\ppm1.00e-5 & 54.73\ppm1.00e-5 & 0.1188\ppm1.90e-8 & -1.63 \\  
        NC-TTT~\cite{osowiechi2024nc} & 48.17\ppm7.00e+0 & 1.086\ppm5.20e-2 & 48.32\ppm1.23e+0 & 0.1440\ppm2.80e-5 & +23.40 & 42.81\ppm1.10e-1 & 40.65\ppm1.10e-1 & 0.1184\ppm7.70e-6 & +31.37 \\  \hline \rowcolor{gray!10}
        S4T (ours) & 59.37\ppm1.52e-1 & 1.052\ppm7.50e-3 & 45.33\ppm7.20e-2 & 0.1441\ppm5.10e-5 &\cellcolor[HTML]{C9D6F7}\textbf{+34.94} & 45.42\ppm1.90e-1 & 41.41\ppm6.30e-1 & 0.1183\ppm5.00e-5 &\cellcolor[HTML]{C9D6F7}\textbf{+34.20} \\ 
    \hline
    \end{tabular}}
    \label{tab:combined_resnet_from_taskonomy}
    % \vspace{-5pt}
\end{table*}

%%%%%%%%%%%%%%%%%%%%%%%%%%%%%%%%%%%%%%%%%%%%%
\begin{figure*}[t]
    \centering
    \includegraphics[width=0.90\linewidth]{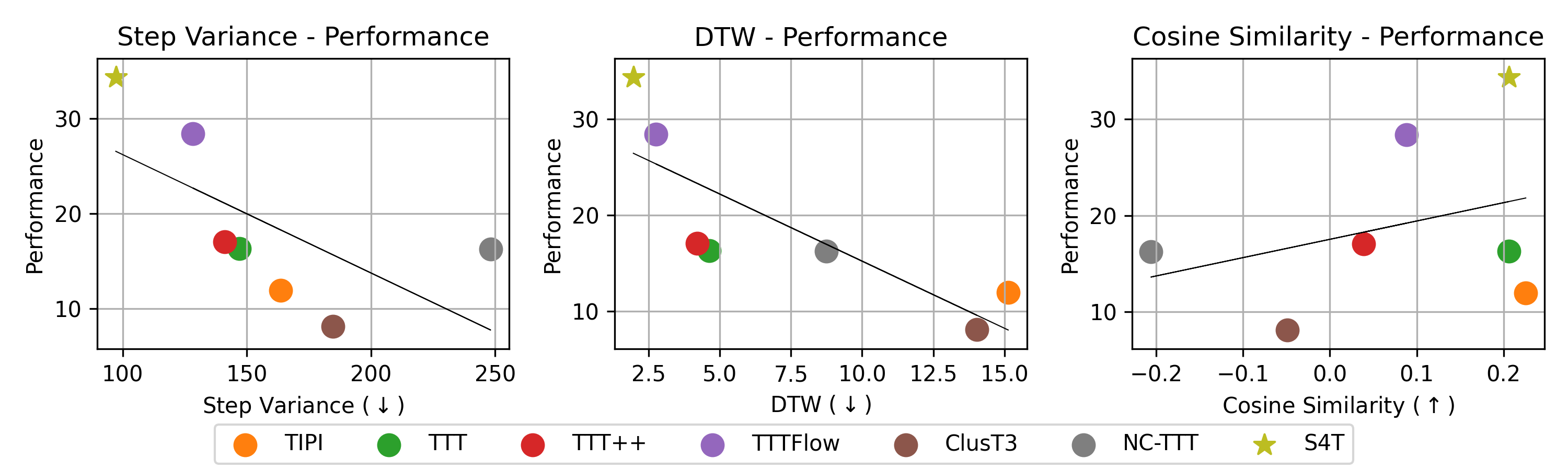}
    \vspace{-1pt}
    \caption{Comparison of previous TTA and TTT methods with proposed S4T in terms of task synchronization. We evaluate the degree of synchronization using various metrics, including step variance, Dynamic Time Warping (DTW), and cosine similarity of task performance (from left to right). S4T achieves high performance across multiple tasks while maintaining strong synchronization.}
    \label{fig:graph_analysis}
\end{figure*}
%%%%%%%%%%%%%%%%%%%%%%%%%%%%%%%%%%%%%%%%%%%%%

%%%%%%%%%%%%%%%%%%%%%%%%%%%%%%%%%%%%%%%%%%%%%%%%%%%%%%%%%%%%%%
\begin{figure*}[t]
    \centering
    \includegraphics[width=0.90\linewidth]{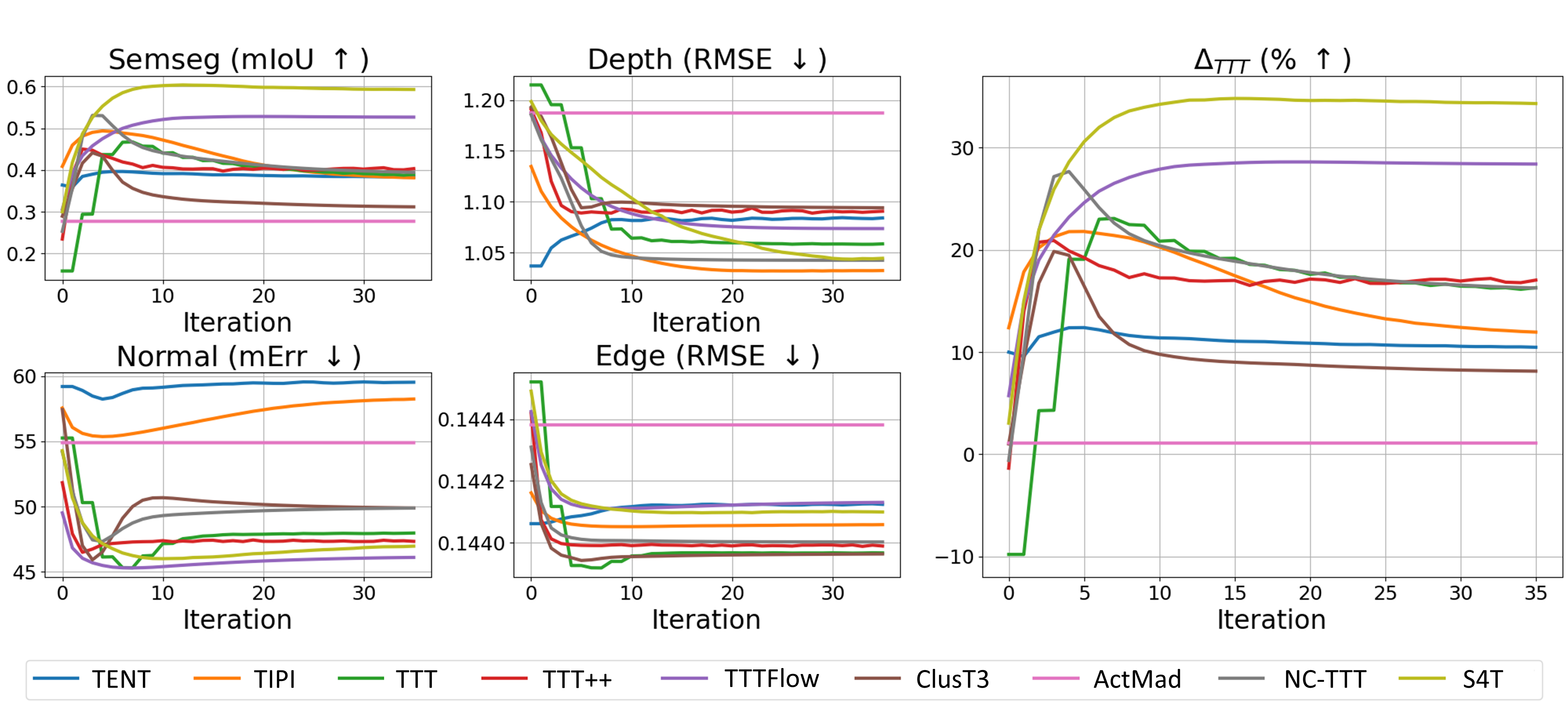}
    \vspace{-5pt}
    \caption{Comparison of previous TTA and TTT methods with our S4T across time steps during test-time training. We evaluate the performance of each task and the overall TTT performance, $\triangle_{TTT}$, under the domain shift from Taskonomy to NYUD-v2.}
    \label{fig:iter}
\end{figure*}
%%%%%%%%%%%%%%%%%%%%%%%%%%%%%%%%%%%%%%%%%%%%%%%%%%%%%%%%%%%%%%

\section{Experiments}
\subsection{Experimental Settings}
\textbf{Datasets.} 
To evaluate the ability to reduce the domain gap on multiple tasks, which include both classification and regression, we utilize several multi-task benchmarks. 
We incorporate NYUD-v2 \cite{RN15}, PASCAL-Context \cite{mottaghi2014role}, and Taskonomy~\cite{RN27} in our TTT evaluation protocols.
These datasets contain 4, 5, and 26 vision tasks, respectively. 
Following the typical protocol for TTT experiments, we use the shared task set between each pair of benchmarks, such as depth estimation, semantic segmentation, normal estimation, and edge detection for NYUD-v2 and Taskonomy. 
For PASCAL-Context and Taskonomy datasets, we use semantic segmentation, normal estimation, and edge detection. 
We select commonly used semantic labels for each dataset pair.
Further details are provided in \cref{append:experimental_details}.

\vspace{2pt}\noindent
\textbf{Baselines and Evaluation Protocols.} We compare our methods with previous test-time adaptation approaches, including TENT \cite{wang2020tent} and TIPI \cite{nguyen2023tipi}, as well as test-time training methods such as TTT \cite{sun2020test}, TTT++ \cite{liu2021ttt++}, TTTFlow \cite{osowiechi2023tttflow}, ClusT3 \cite{hakim2023clust3}, ActMAD \cite{mirza2023actmad}, and NC-TTT~\cite{osowiechi2024nc}. 
To evaluate the adaptation, we cover the domain shifts as follows: 1) Taskonomy$\rightarrow$NYUD-v2, 2) Taskonomy$\rightarrow$PASCAL-Context in the main paper, and 3) NYUD-v2$\rightarrow$Taskonomy, 4) PASCAL-Context$\rightarrow$Taskonomy in \cref{append:exp}.
For a fair comparison across various TTT methods, we use identical multi-task architectures. Specifically, we adopt a single backbone with multiple task-specific decoders, which is the most commonly used baseline in the MTL domain~\cite{riquelme2021scaling, zhang2022mixture, fan2022m3vit, mustafa2022multimodal, chen2023mod, huang2024going}. 
During test-time, we evaluate all tasks simultaneously. 
For TTA, we adapt the model that was trained on multiple tasks in the source domain. 
To evaluate overall improvements during test time, we propose a metric, $\triangle_{TTT}$ for assessing TTT, motivated by \cite{RN2}. 
This metric measures averaged per-task performance improvements and is defined as: $\triangle_{TTT} = \frac{1}{n} \sum_{i=1}^{n} (-1)^{l_i} \frac{M_{TTT,i} - M_{b,i}}{M_{b,i}}$. In this equation, \(M_{TTT,i}\) indicates the performance of task \(i\) when TTT is applied, while \(M_{b,i}\) represents the performance of task \(i\) without TTT. The value \(l_i = 1\) if a lower measure \(M_i\) indicates better performance for task \(i\), and \(l_i = 0\) otherwise.

\vspace{2pt}\noindent
\textbf{Implementation Details.} For experiments, we use ResNet50 as an encoder and task-specific decoders combine multi-layer features with convolutional layers. 
A single convolutional layer is used for task-specific projection, and a lightweight vision transformer is employed for TBS, which increases the network size by $24.1\%$ when applied to ResNet50. The models are trained for 40,000 iterations on the source domain with a batch size of 8, followed by sequential training on the target domain. 
We utilize the loss scales and functions commonly used in MTL literature \cite{yang2024multi,ye2022taskprompter,invpt,vandenhende2020mti,zhang2019pattern}. 
More details are provided in \cref{append:experimental_details}.

\subsection{Experimental Results}
\textbf{Comparison with Previous Methods.} We compare S4T with previous state-of-the-art TTT methods. 
Taskonomy is used as the source domain, and results for NYUD-v2 and Pascal-Context as target domains are presented in \cref{tab:combined_resnet_from_taskonomy}. 
Since each method converges at different rates, we select the point at which each method achieves its best TTT performance, averaged across all tasks, measured by $\triangle_{TTT}$ for a fair comparison. 
S4T outperforms all other methods in both settings. 
A key observation is that the effectiveness of previous methods depends on the type of main task. 
Methods like TENT and NC-TTT, which rely on class-level clustering to reduce the domain gap, exhibit limited performance on regression tasks across both datasets. 
Even in classification tasks such as semantic segmentation, feature-level adaptation methods that use class cluster information, like ClusT3, ActMad, and NC-TTT, show limited effectiveness. 
This suggests that these methods are better suited for simple classification tasks and struggle to generalize to more complex dense prediction tasks.
The results also indicate that TTT performance heavily relies on the choice of unsupervised tasks selected for auxiliary training. 
In contrast, our S4T method captures task relations and effectively incorporates them into the adaptation process, achieving superior performance across multiple tasks.  

\vspace{2pt}\noindent
\textbf{Degree of Synchronization.} In \cref{fig:graph_analysis}, we evaluate the degree of synchronization of previous TTA and TTT methods with proposed S4T. To assess synchronization across tasks, we utilize three key metrics: Step Variance, Dynamic Time Warping, and Cosine Similarity. (1) Step Variance (SV) measures the dispersion in peak adaptation steps across tasks. Lower variance indicates synchronized adaptation, while higher variance suggests misalignment. (2) Dynamic Time Warping (DTW) quantifies the alignment of adaptation trajectories. A lower DTW implies similar adaptation patterns, whereas a higher value indicates divergence in task behaviors. (3) Cosine Similarity (CS) evaluates the directional consistency of performance changes. A higher CS suggests strong synchronization in adaptation trends, while a lower value indicates inconsistency. Detailed metrics are introduced in \cref{append:sync_metric}. In \cref{fig:graph_analysis}, we observe a strong correlation between synchronization and performance improvement ($\triangle_{TTT}$). Methods achieving lower SV, lower DTW, and higher CS tend to exhibit better adaptation performance. Notably, S4T enhances synchronization across tasks, aligning adaptation trajectories and improving consistency in multi-task adaptation.

\vspace{2pt}\noindent
\textbf{Performance Over Time with Adaptation Iterations.}
We evaluate the performance of each TTT method in an online manner over time with adaptation iterations. As shown in Fig.~\ref{fig:iter}, the performance of S4T continuously improves with an increasing number of time steps. In contrast, most other adaptation methods experience performance degradation during longer adaptation processes. This phenomenon has been frequently reported in previous research, such as TENT, which indicated that adaptation loss has a detrimental influence on learning the target task over longer adaptation periods. This is a crucial point in practice since we often do not know how many adaptation steps are needed during test-time. S4T is less affected by this issue because it directly leverages the relations between the main tasks that the network is trying to adapt. Additionally, in situations where multiple tasks need to be adapted across domains, S4T offers more advantages as it avoids problems related to differing convergence rates between tasks.

\begin{table*}[t!]
\vspace{-10pt}
  \centering
  \caption{Ablation study for individual components of S4T.}
  \vspace{-5pt}
    \resizebox{0.85\linewidth}{!}
    {
    \begin{tabular}{c|ccccc|ccccc}
    \toprule
    Benchmark 
    & \multicolumn{5}{c}{Taskonomy $\rightarrow$ NYUD-v2} 
    & \multicolumn{5}{c}{Taskonomy $\rightarrow$ PASCAL-Context}\\
    
    \cmidrule(lr){1-1} \cmidrule(lr){2-6} \cmidrule(lr){7-11}
    % \midrule
    
    TBS 
    & - & \checkmark & \checkmark& \checkmark & \checkmark 
    & - & \checkmark & \checkmark& \checkmark & \checkmark \\
    
    Task-specific Projection
    & - & - &\checkmark & - & \checkmark
    & - & - &\checkmark & - & \checkmark\\

    Feat. Masking
    & - & - & - & \checkmark& \checkmark 
    & - & - & - & \checkmark& \checkmark \\

    Image Recon. Task
    & - & - & - & \checkmark & - 
    & - & - & - & \checkmark & - \\
    
    $\triangle_{TTT} \uparrow (\%)$
    & +0.00  & +26.16  & +33.79 & +26.76  & +34.94
    & +0.00  & +28.32  & +32.27 & +26.43  & +34.20\\
    \bottomrule
    \end{tabular}%
    }
  \label{tab:ablation}%
    \vspace{-5pt}
\end{table*}%
%%%%%%%%%%%%%%%%%%%%%%%%%%%%%%%%%%%%%%%%%%%%%%%%%%%%%%%%%%%%%%

\subsection{Ablation Study}
In this section, we present additional ablation experiments to evaluate each component of S4T and their respective strategies. We assess the influence of the following components: (1) the  Task Behavior Synchronizer (TBS), including the joint task prediction loss $\mathcal{L}^{TBS}$, (2) task-specific projection, including $\mathcal{L}^{TP}$, (3) feature masking applied to each task-specific latent vector, and (4) a comparison of results when using image reconstruction as an auxiliary task instead of the main tasks we aim to adapt for the TTT branch. In \cref{tab:ablation}, we show the performance improvements in TTT based on different combinations of these components, with improvements measured relative to results without any TTT methods, denoted as $\triangle_{TTT} \uparrow (\%)$.

%%%%%%%%%%%%%%%%%%%%%%%%%%%%%%%%%%%%%%%%%%%%%%%%%%%%%%%%%%%%%%

\begin{table*}[t!]
\centering
\caption{Ablation study on the impact of different masking strategies, as described in Fig.~\ref{fig:mask}.}
\vspace{-5pt}
\small
\begin{tabular}{c|cccc}
\hline
Method                                &(a) Random & (b) Not Overlap & (c) Same for All & (d) Hide specific Tasks \\ \hline
$\triangle_{TTT} \uparrow (\%)$       & +32.76       & +32.56            & +34.94             & +33.06                    \\ \hline
\end{tabular}
\label{tab:mask_strategy}
\vspace{-8pt}
\end{table*}

\vspace{2pt}\noindent
\textbf{Ablation on Each Component.} Using the TBS alone effectively reduces the domain gap, leading to performance improvements of 26.16\% for NYUD-v2 and 28.32\% for PASCAL-Context.
In this scenario, the TBS captures task relations by leveraging shared representations across multiple tasks instead of using extracted task-specific latent vectors. When task-specific projections are introduced to obtain task-specific latent vectors for TBS, performance further improves. This suggests that task relations are more effectively captured and utilized for synchronizing task behavior during adaptation when distinct task-specific information is incorporated.
Additionally, the feature masking strategy provides further performance gains, although most of the improvements are driven by the TBS and task-specific projection. Lastly, we integrate an image reconstruction task into the TTT branch to assess its impact. In this setup, TBS predicts the reconstructed image instead of task labels. The learned information from image reconstruction resulted in significantly poorer TTT performance compared to our approach. This highlights that using auxiliary tasks, such as image reconstruction, does not necessarily ensure the inclusion of useful information for downstream tasks.

%%%%%%%%%%%%%%%%%%%%%%%%%%%%%%%%%%%%%%%%%%%%%%%%%%%%%%%%%%%%%%
\begin{figure}[t]
    \centering
    \includegraphics[width=0.99\linewidth]{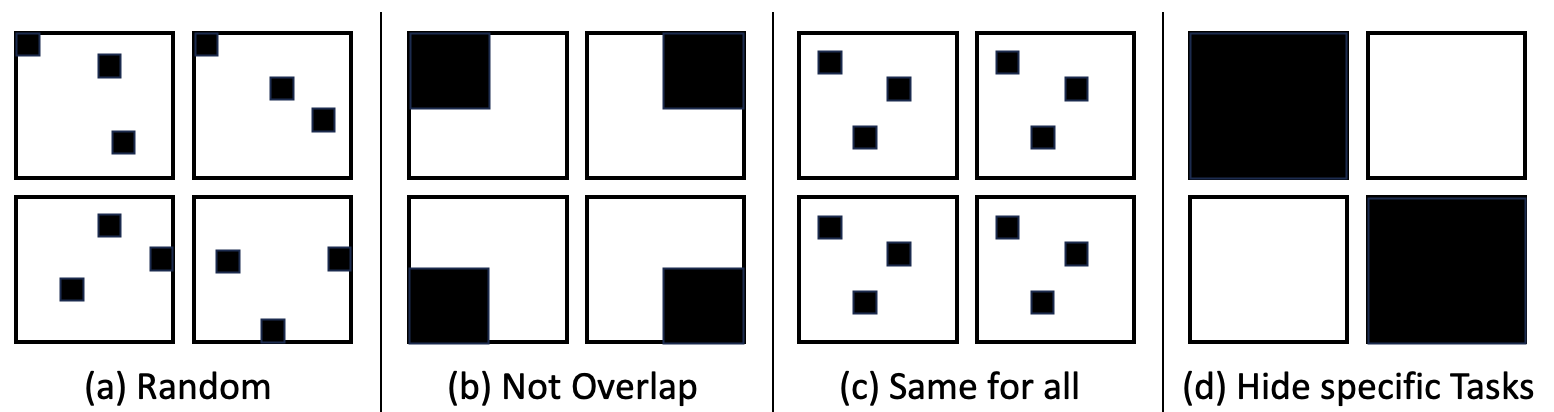}
    \vspace{-4pt}
    \caption{Candidates for masking strategy $\mathcal{M}_i$ applied to task-specific latent vectors, $\tilde{z}_i = \mathcal{M}_i(z_i)$. Each task-specific latent vector $z_i$ is represented by separate large squares, with the unmasked portions shaded in black. 
    In (a), randomly select patches for masking. In (b), mask without overlap between tasks. In (c), apply the same masking strategy across all task-specific latent vectors. In (d), completely mask the latent vector of a specific task.
    }
    \label{fig:mask}
    \vspace{-8pt}
\end{figure}
%%%%%%%%%%%%%%%%%%%%%%%%%%%%%%%%%%%%%%%%%%%%%%%%%%%%%%%%%

\vspace{2pt}\noindent
\textbf{Masking Strategy and Masking Ratio.} To evaluate which masking strategy $\mathcal{M}$ for task-specific latent vectors $\tilde{z}_i = \mathcal{M}_i (z_i)$ would be beneficial for learning inter-task relations, we select several candidates for masking strategies to assess their influence, as shown in Fig.~\ref{fig:mask}. We consider four scenarios: (a) we randomly mask each task-specific latent vector $z_i$, (b) we mask them without overlap across tasks, (c) we mask identical patches, which are randomly chosen for all tasks, and (d) we randomly select task sets and entirely mask their task-specific latent vectors. As shown in \cref{tab:mask_strategy}, (c) Same for All shows the best performance, thus we adopt this strategy for our methods. We guess there are two reasons why (c) produces the best performance compared to the other strategies. First, although the task-specific latent vector $z_i$ is derived from a task-specific projection, it may still contain shared representations from other tasks. In such cases, using each task's latent vector for prediction results in trivial predictions by the TBS.
Second, predicting the task label for a masked patch from the unmasked patch encourages the TBS to capture spatially global information across different task-specific latent vectors. If the TBS has access to the same patch location from another task's latent vector, it might merely memorize the style transfer between these vectors, which would negatively affect generalization.
In \cref{append:exp}, we evaluate the influence of the masking ratio for the adopted masking strategy (c) on the performance of tasks during test-time. The overall TTT performance improves as the masking ratio increases, peaking at approximately 0.7 to 0.8. It is noteworthy that the overall trend is quite consistent across tasks.

Additional experiments on different domain shift scenarios, the effect of using a single task for adaptation, and inference time are provided in \cref{append:exp}.

%% file: sec/5_conclusion.tex
\section{Conclusion}
In this paper, we introduce the unsynchronization problem, which arises when conventional TTT methods are applied to adapt multiple tasks in the target domain. Unlike conventional TTT approaches that rely on independent auxiliary tasks, we propose S4T, a novel approach that leverages task relations to synchronize adaptation across multiple tasks. By incorporating the Task Behavior Synchronizer, our method ensures that adaptation steps are aligned across different tasks, mitigating the performance misalignment observed in previous methods and enhancing overall adaptation consistency. Through our analysis, we demonstrate that stronger synchronization leads to improved post-adaptation performance, highlighting the importance of task relations in TTT. Extensive experiments on multi-task benchmarks validate the effectiveness of our approach.

%% file: sec/append.tex
\clearpage
\appendix
\maketitlesupplementary

\section{Additional Experimental Details}
\label{append:experimental_details}
\textbf{Experimental Settings.} In the training phase within the source domain, we utilize the Adam optimizer \cite{kingma2014adam} with a polynomial decay for the learning rate. We set the learning rate to \(2 \times 10^{-5}\) and the weight decay to \(1 \times 10^{-6}\) for training the networks. The batch size is 8, and we perform 60,000 iterations for training. We set $\lambda^{TP}$=$\lambda^{TBS}$=$1$, and reduce $\lambda^{TBS}$ to 0.01 at the test time. During test time, we adopt the SGD optimizer to ensure stable convergence with the TTT loss. The learning rate remains the same. During test time, we update the network for each batch of data for up to 40 steps in an online manner.

%%%%%%%%%%%%%%%%%%%%%%%%%%%%%%%
\begin{table}[h]
\caption{Hyperparameters for experiments.}
\centering
\renewcommand\arraystretch{1.20}
\begin{tabular}{lc}
\hline
Hyperparameter                  &  Value \\ \hline
$\llcorner$ Scheduler                       &  Polynomial Decay\\
$\llcorner$ Minibatch size                  &  8\\
$\llcorner$ Backbone                        &  ResNet50 \cite{he2016deep}  \\
\hspace{10pt}$\llcorner$ Learning rate                   &  0.00002\\
\hspace{10pt}$\llcorner$ Weight Decay                    &  0.000001\\ \hline
Train Time Training             & \\
$\llcorner$ Optimizer                       &  Adam \cite{kingma2014adam}\\
$\llcorner$ Number of iterations            &  60000\\
\hspace{10pt}$\llcorner$ Learning rate                   &  0.00002\\
\hspace{10pt}$\llcorner$ Weight Decay                    &  0.000001\\
\hline
Test Time Training \\
$\llcorner$ Optimizer                       &  SGD \\
$\llcorner$ Minibatch size                  &  8\\
$\llcorner$ Number of steps                 &  40\\
\hline
\end{tabular}
\label{Implementation_details}
\end{table}
%%%%%%%%%%%%%%%%%%%%%%%%%%%%%%%

\vspace{2pt}
\noindent\textbf{Metrics.} For semantic segmentation, we utilize the mean Intersection over Union (mIoU) metric. The performance of surface normal prediction was measured by calculating the mean angle distances between the predicted output and the ground truth. To evaluate depth estimation and edge detection, we use the Root Mean Squared Error (RMSE).

\vspace{2pt}
\noindent\textbf{Datasets.} To implement TTT in semantic segmentation tasks on different datasets (Taskonomy $\leftrightarrow$ NYUD-v2, Taskonomy $\leftrightarrow$ PASCAL-Context), we find shared class labels in each of the two datasets.
For Taskonomy $\leftrightarrow$ NYUD-v2, we use 6 shared classes: \texttt{table, tv, toilet, sofa, potted plant, chair}.
For Taskonomy $\leftrightarrow$ PASCAL-Context, we use 7 class labels: \texttt{refrigerator, table, toilet, sofa, bed, sink, chair}.
We use the split of train/test following the common multi-task benchmarks, NYUD-v2, PASCAL-Context and Taskonomy.
In the case of NYUD-v2, we utilize 795 images for training and reserve 654 images for test-time training.
With PASCAL-Context, 4,998 images are employed during training, and 5,105 images are used for test-time training.
For Taskonomy, we leverage 295,521 images for training and apply 5,451 images during test-time.

\section{Evaluation of Synchronization}
\label{append:sync_metric}
We qualify the limitations of the existing TTT methods on multi-task settings through three points: step variance (SV) of peak step during adaptation, dynamic time warping (DTW) and cosine similarity (CS) of adaptation graph. We denote that the steps $\tau \in \{1,2,\dots,\mathrm{T} \}$ and $y_i^{(\tau)}$ denotes the performance of $i$-th task at step $\tau$.
$n$ is the number of tasks.
Step Variance measures the dispersion of the steps where tasks achieve their maximum performance:
\begin{align}
    \text{SV} = \sqrt{\frac{1}{n} \sum_{i=1}^n \left( \tau_i^{peak} - \bar{\tau}^{peak} \right)^2}.
\end{align}
where $\tau_{i}^{peak}$ denotes the step with best performance for each task $i$ and $\bar{\tau}^{peak} = \frac{1}{n} \sum_{i=1}^n \tau_i^{peak}$.

DTW aligns the performance trajectories \( Y_i \) and \( Y_j \) of the different tasks by computing the optimal alignment path:
\begin{equation}
\begin{split}
\text{DTW} &= \frac{1}{C(n, r)} \sum_{i=1}^{n} \sum_{j>i}^n \text{DC}_{i,j}(T,T), \\
\text{DC}_{i,j}(\tau,\kappa) &= d(y_i^{(\tau)}, y_j^{(\kappa)}) + \min \left\{
\begin{array}{l}
\text{DC}_{i,j}(\tau-1, \kappa), \\
\text{DC}_{i,j}(\tau, \kappa-1), \\
\text{DC}_{i,j}(\tau-1, \kappa-1)
\end{array}
\right.
\end{split}
\end{equation}
where \( d \) represents a distance metric, with the Euclidean distance used in our case, and \( C \) denotes the binomial coefficient, representing the number of combinations.

Cosine Similarity evaluates the directional consistency of performance changes between two tasks:
\begin{align}
    \text{Cosine Similarity} &= \frac{1}{C(n, r)} \sum_{i=1}^{n} \sum_{j>i}^n \sum_{\tau=1}^{T-1} \text{CS}_{i,j}(\tau), \\
    \text{CS}_{i,j}(\tau) &= \frac{\Delta y_i^{\tau} \cdot \Delta y_j^{\tau}}{\|\Delta y_i^{\tau}\| \|\Delta y_j^{\tau}\|},
\end{align}
where \( \Delta y_i^{\tau} = y_i^{\tau+1} - y_i^{\tau} \), and \( C \) represents the binomial coefficient.

\section{Reliability of Assumptions}
\label{sup:reliability}

\begin{figure}[h]
    \centering
    \vspace{-5pt}
    \includegraphics[width=0.90\linewidth]{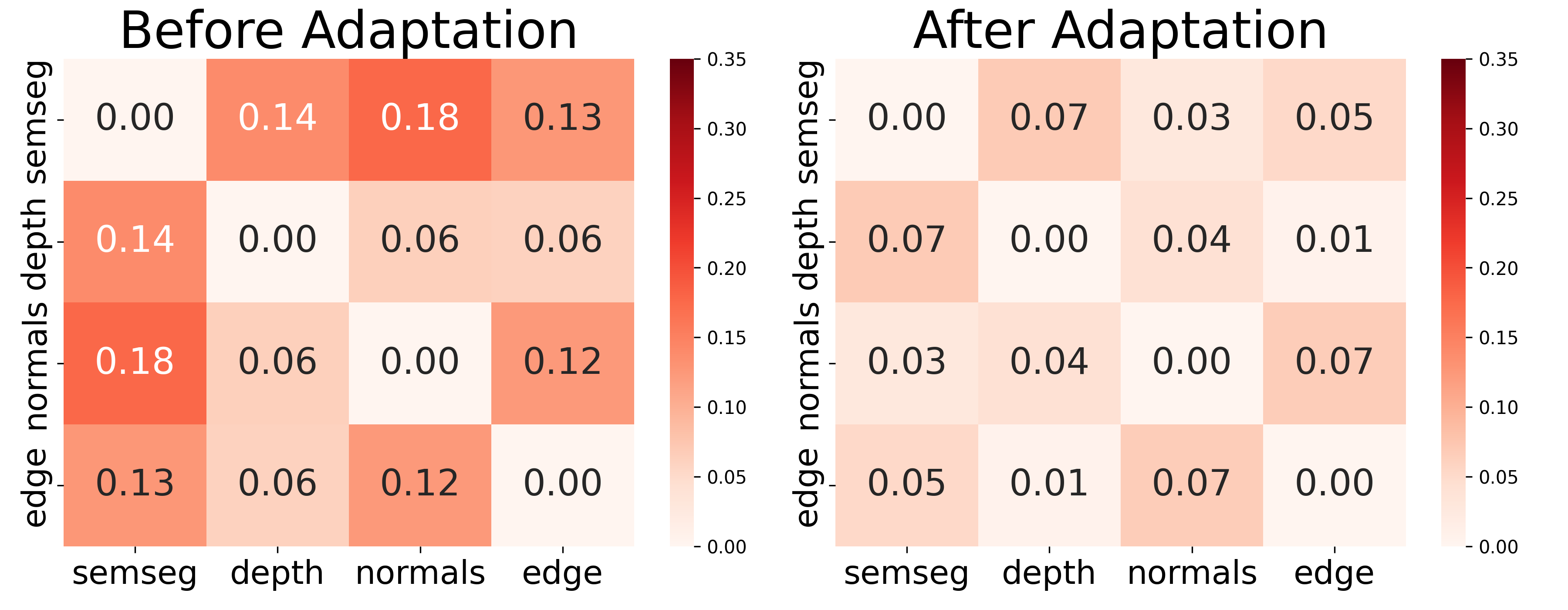}
    \vspace{-5pt}
    \caption{Task affinity gap before and after adaptation.}
    \label{fig:assum1}
    \vspace{-5pt}
\end{figure}

\noindent\textbf{Reliability of Assumption \textcolor{red}{1}.}
Assumption \textcolor{red}{1} posits that task relations in the target domain can be reliably captured by learning a domain-dependent transformation function $f$ during test-time training. To support this, we compute task affinity matrices using the cosine similarity between task-specific latents learned in the source domain. We compare these with the affinity matrices obtained from the fine-tuned model on the target domain, plotting the gap in \cref{fig:assum1}, which shows the domain shift from Taskonomy to NYUD-v2. While the initial task relations differ across domains (\textit{Before Adaptation}), the learned transformation $f$ effectively reduces the affinity gap (\textit{After Adaptation}), supporting the validity of the assumption in real-world scenarios.

%%%%%%%%
%%%%%%%%

\begin{figure}
    \centering
    \vspace{-5pt}
    \includegraphics[width=0.80\linewidth]{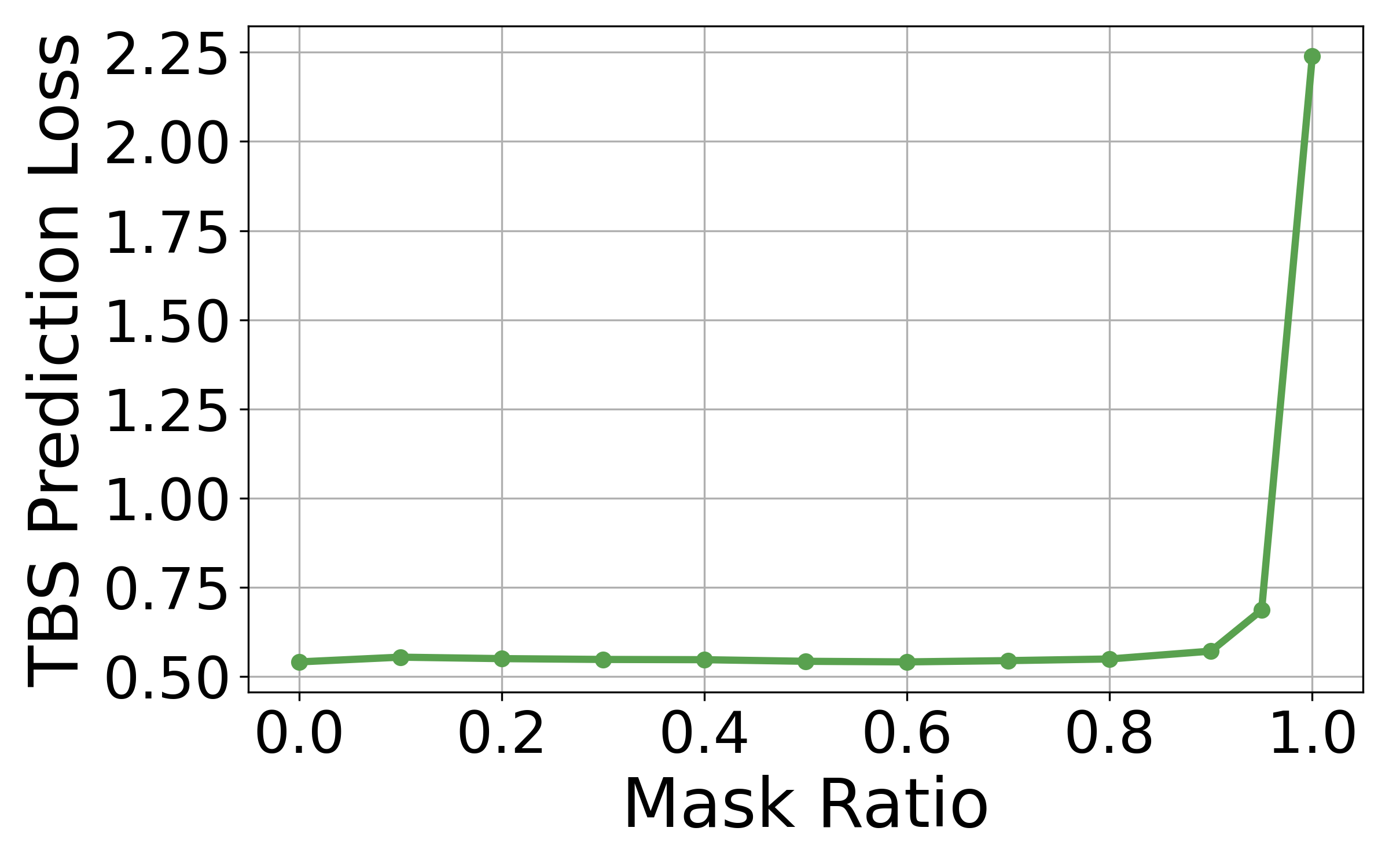}
    \vspace{-8pt}
    \caption{TBS prediction loss across varying masking ratios.}
    \label{fig:assum2}
    \vspace{-5pt}
\end{figure}

\noindent
\textbf{Reliability of Assumption \textcolor{red}{2}.}
Assumption~\textcolor{red}{2} does not claim that TBS achieves perfect predictions of task labels, but rather that its predictions from masked latents $\tilde{z}$ are \emph{comparable} to those from unmasked latents $z$ once sufficiently trained. We evaluate this by directly measuring the TBS prediction loss $\mathcal{L}^{TBS}$ under varying mask ratios. As shown in \cref{fig:assum2}, TBS maintains over 90\% of its original performance at $r = 0$ even when the mask ratio increases to $r = 0.9$, supporting the validity of the assumption in practice. 
This assumption is not directly comparable to those in image-level restoration settings like MAE, as TBS predicts task labels from masked latent features, rather than reconstructing pixel-level content.

\noindent
Additionally, to quantify the level of distribution shift under which our assumptions hold, we evaluate $\triangle_{TTT}$ by adding Gaussian noise to input images, where the noise standard deviation is set to $\alpha \cdot \sigma_{\text{img}}$ ($\alpha$: noise scale), with $\sigma_{\text{img}}$ being the standard deviation of input images. 
As shown in \cref{tab:noise}, the results suggest that our method and proposition hold robustly under moderate shifts.
%%%%%%%%%%%%%%
\begin{table}[h]
\scriptsize
\vspace{-5pt}
\centering
\caption{$\triangle_{TTT}$ under Gaussian noise w.r.t. noise scale $\alpha$}
\vspace{-5pt}
\begin{tabular}{c|cccccccc} \hline
$\alpha$                        &0      &0.01   &0.05   &0.1    &0.2    &0.3    &0.4  \\ \hline
$\triangle_{TTT} \uparrow$      &34.89  &34.42  &33.80  &32.15  &28.64  &23.80  &19.73     \\ \hline
\end{tabular}
\vspace{-10pt}
\label{tab:noise}
\end{table}
%%%%%%%%%%%%%%

\section{Additional Experiments and Analyses}
\label{append:exp}

\vspace{2pt}
\noindent\textbf{Comparison with Previous Methods in Different Scenarios.} We compare S4T with previous state-of-the-art TTT methods in different scenarios, using NYUD-v2 and PASCAL-Context as the source domains and Taskonomy as the target domain. The results are presented in \cref{tab:combined_resnet_to_taskonomy}. For a fair comparison, we select the point at which each method achieves its best TTT performance, averaged across all tasks, as measured by $\triangle_{TTT}$. Since NYUD-v2 and PASCAL-Context have smaller datasets, the overall TTT performance is lower compared to scenarios where Taskonomy is used as the source domain. The proposed S4T still demonstrates comparable performance in these scenarios.

\vspace{2pt}
\noindent\textbf{Evaluation of S4T Using Single Task for Adaptation.} 
To ensure that S4T’s gains are not solely due to multi-task structures, we evaluated it in a single-task setting where only one task label is available during adaptation. In this case, TBS uses a single latent vector per task. As shown in \cref{tab:combined_resnet_single}, performance drops compared to the multi-task setup, highlighting the importance of modeling task relations. Still, S4T achieves competitive results, confirming that its benefits stem from the core TTT mechanism rather than architectural scale or multi-head design.

\vspace{2pt}
\noindent\textbf{Number of Parameters and Inference Time.} As shown in Table~\ref{tab:params_inference_time}, we compare the number of parameters and inference time from Taskonomy to NYUD-v2 for S4T, against previous SOTA TTT methods. For inference, we use NVIDIA RTX A6000 48GB with batch size 8.

\vspace{2pt}
\noindent\textbf{Comparison with a Strong MTL+TTT Baseline.} We compare S4T with a strong TTT baseline that combines an existing TTT method (NC-TTT~\cite{sun2020test}) and a high-performing multi-task learning MTL architecture (MTI-Net~\cite{vandenhende2020mti}). As shown in \cref{tab:mti}, S4T achieves slightly higher adaptation gain ($\triangle_{\text{TTT}} = 34.9$ vs. 32.5) while using fewer than 20\% of the parameters (29.2M vs. 165.0M) and maintaining comparable inference time (0.3657s vs. 0.3208s). This indicates that our method is not only effective in terms of performance but also significantly more parameter-efficient. Importantly, S4T is designed as a modular plug-in that can be ad-hoc attached to existing MTL architectures.

%%%%%%%%%%%%%%%%%%%
\begin{table}[h]
\caption{Comparison of S4T and NC-TTT + MTI-Net}
\vspace{-5pt}
\centering
\resizebox{\linewidth}{!}
{
\begin{tabular}{l|cccc}
\hline
Method              &  $\triangle_{TTT} \uparrow$&  Params (M)& Inference Time (s) \\ \hline
NC-TTT + MTI-Net    &32.5  &165.0  &0.3208  \\
Ours                &34.9  &29.2  &0.3657  \\ \hline
\end{tabular}
\label{tab:mti}
}
\vspace{-5pt}
\end{table}
%%%%%%%%%%%%%%%%%%%

\begin{table*}[ht]
    \centering
    \renewcommand{\arraystretch}{1.30}
    \caption{Comparison of multi-task performance from NYUD-v2 to Taskonomy across different tasks for S4T, against previous TTA and TTT methods.}
    \resizebox{\linewidth}{!}
    {
    \begin{tabular}{l|cccc|>{\columncolor[HTML]{C9D6F7}}c|ccc|>{\columncolor[HTML]{C9D6F7}}c} 
    \hline \rowcolor{gray!30}
        Dataset & \multicolumn{5}{c|}{NYUD-v2 $\rightarrow$ Taskonomy} & \multicolumn{4}{c}{PASCAL-Context $\rightarrow$ Taskonomy} \\ \hline \rowcolor{gray!30}
        Task & \multicolumn{1}{c}{Semseg} & \multicolumn{1}{c}{Depth} & \multicolumn{1}{c}{Normal} & \multicolumn{1}{c}{Edge} & \multicolumn{1}{c|}{$\triangle_{TTT}$} & \multicolumn{1}{c}{Semseg} & \multicolumn{1}{c}{Normal} & \multicolumn{1}{c}{Edge} & \multicolumn{1}{c}{$\triangle_{TTT}$} \\ \rowcolor{gray!30}
        Metric & mIoU $\uparrow$ & RMSE $\downarrow$ & mErr $\downarrow$ & RMSE $\downarrow$ & $\%$ $\uparrow$ & mIoU $\uparrow$ & mErr $\downarrow$ & RMSE $\downarrow$ & $\%$ $\uparrow$ \\ \hline
        Base & 48.21\ppm1.580e-2 & 0.0507\ppm2.000e-4 & 27.60\ppm1.200e-2 & 0.3058\ppm2.000e-4 & +0.00 & 50.94\ppm6.630e-1 & 31.27\ppm7.100e-2 & 0.3032\ppm1.000e-4 & 0.00 \\ \hline
        TENT~\cite{wang2020tent} & 39.75\ppm1.200e-2 & 0.0634\ppm0.000e+0 & 37.49\ppm3.500e-1 & 0.3084\ppm5.000e-4 & -19.76 & 44.68\ppm3.530e-1 & 42.32\ppm1.830e-1 & 0.3269\ppm2.100e-4 & -0.18 \\  
        TIPI~\cite{nguyen2023tipi} & 47.03\ppm7.080e-4 & 0.0514\ppm3.550e-8 & 28.40\ppm1.490e-5 & 0.3052\ppm3.320e-8 & -1.64 & 51.48\ppm1.200e-3 & 32.33\ppm4.180e-5 & 0.3031\ppm1.490e-7 & -0.77 \\  \hline
        TTT~\cite{sun2020test} & 49.66\ppm4.120e-1 & 0.0523\ppm4.100e-3 & 31.96\ppm1.190e-1 & 0.3094\ppm6.000e-4 & -4.28 & 48.00\ppm2.661e+0 & 37.77\ppm2.214e+0 & 0.3048\ppm3.600e-4 & -9.00 \\  
        TTT++~\cite{liu2021ttt++} & 39.29\ppm1.089e-1 & 0.0595\ppm1.300e-3 & 36.53\ppm4.160e-1 & 0.3132\ppm2.610e-3 & -17.65 & 38.66\ppm3.090e-1 & 39.81\ppm3.850e-1 & 0.3050\ppm6.480e-4 & -17.33 \\  
        TTTFlow~\cite{osowiechi2023tttflow} & 48.56\ppm3.000e-1 & 0.0540\ppm1.000e-4 & 34.36\ppm1.425e-1 & 0.3086\ppm1.000e-4 & -7.73 & 51.55\ppm3.950e-1 & 34.60\ppm1.200e-1 & 0.3042\ppm1.500e-3 & -3.26 \\ 
        ClusT3~\cite{hakim2023clust3} & 51.14\ppm1.757e+0 & 0.0516\ppm3.700e-4 & 30.03\ppm4.310e-1 & 0.3065\ppm5.000e-4 & -1.16 & 49.67\ppm6.480e-1 & 35.22\ppm1.570e-1 & 0.3019\ppm1.900e-4 & -4.90 \\ 
        ActMAD~\cite{mirza2023actmad} & 55.04\ppm7.300e-4 & 0.0506\ppm5.800e-8 & 27.88\ppm7.000e-5 & 0.3081\ppm1.100e-9 & +3.17 & 51.79\ppm8.030e-1 & 31.10\ppm1.240e-1 & 0.3031\ppm1.020e-4 & +0.74 \\  
        NC-TTT~\cite{osowiechi2024nc} & 49.95\ppm6.530e-1 & 0.0516\ppm4.600e-5 & 29.95\ppm4.200e-2 & 0.3093\ppm1.010e-4 & -1.96 & 48.78\ppm5.100e-1 & 32.86\ppm2.220e+0 & 0.3040\ppm1.300e-3 & -3.19 \\  \hline \rowcolor{gray!10}
        S4T (ours) & 53.12\ppm1.340e-1 & 0.0511\ppm1.930e-4 & 27.58\ppm1.044e-1 & 0.3089\ppm3.540e-5 &\cellcolor[HTML]{C9D6F7}\textbf{+2.13} & 53.18\ppm3.150e-1 & 31.50\ppm7.890e-2 & 0.3036\ppm2.000e-4 &\cellcolor[HTML]{C9D6F7}\textbf{+1.18} \\ 
    \hline
    \end{tabular}}
    \label{tab:combined_resnet_to_taskonomy}
\end{table*}

%%%%%%%%%%%%%%%%%%%%%%%%%%%%%%
\begin{table*}[ht]
    \centering
    \renewcommand{\arraystretch}{1.30}
    \caption{We compare the TTT performance of Taskonomy as the source domain and NYUD-v2 as the target domain across four tasks for S4T, analyzing both single-task and multi-task scenarios.}
    \vspace{-8pt}
    \resizebox{\linewidth}{!}
    {
    \begin{tabular}{l|cccc|>{\columncolor[HTML]{C9D6F7}}c|ccc|>{\columncolor[HTML]{C9D6F7}}c} 
    \hline \rowcolor{gray!30}
        Dataset & \multicolumn{5}{c|}{NYUD-v2 $\rightarrow$ Taskonomy} & \multicolumn{4}{c}{PASCAL-Context $\rightarrow$ Taskonomy} \\ \hline \rowcolor{gray!30}
        Task & \multicolumn{1}{c}{Semseg} & \multicolumn{1}{c}{Depth} & \multicolumn{1}{c}{Normal} & \multicolumn{1}{c}{Edge} & \multicolumn{1}{c|}{$\triangle_{TTT}$} & \multicolumn{1}{c}{Semseg} & \multicolumn{1}{c}{Normal} & \multicolumn{1}{c}{Edge} & \multicolumn{1}{c}{$\triangle_{TTT}$} \\ \rowcolor{gray!30}
        Metric & mIoU $\uparrow$ & RMSE $\downarrow$ & mErr $\downarrow$ & RMSE $\downarrow$ & $\%$ $\uparrow$ & mIoU $\uparrow$ & mErr $\downarrow$ & RMSE $\downarrow$ & $\%$ $\uparrow$ \\ \hline
        Base & 29.31\ppm6.300e-2 & 1.179\ppm8.000e-3 & 61.32\ppm8.200e-1 & 0.1443\ppm7.100e-4 & +0.00 & 27.08\ppm1.400e-2 & 63.46\ppm9.540e-1 & 0.1185\ppm7.100e-4 & 0.00 \\ \hline
        S4T (single) & 56.56\ppm8.500e-2 & 1.065\ppm5.600e-3 & 47.32\ppm3.000e-2 & 0.1444\ppm1.900e-5 & - & 43.28\ppm3.700e-1 & 46.91\ppm7.500e-2 & 0.1185\ppm9.600e-6 & - \\
        S4T (multi) & 59.37\ppm1.520e-1 & 1.052\ppm7.500e-3 & 45.33\ppm7.200e-2 & 0.1441\ppm5.100e-5 & +34.94 & 45.42\ppm1.900e-1 & 41.41\ppm6.300e-1 & 0.1183\ppm5.000e-5 & +34.20 \\
    \hline
    \end{tabular}
    }
\label{tab:combined_resnet_single}
\end{table*}
%%%%%%%%%%%%%%%%%%%%%%%%%%%%%%
\begin{table*}[t!]
\centering
\caption{Comparisons of number of parameters and inference time from Taskonomy to NYUD-v2 for S4T, against previous TTT methods.}
\vspace{-8pt}
\label{tab:params_inference_time}
\small
\begin{tabular}{l|cccccc|c}
\toprule
                     & TTT      & TTT++     & TTTFlow  & ClusT3   & ActMad   & NC-TTT   & S4T(Ours) \\
\midrule
Number of Parameters & 27.49M   & 149.40M   & 49.56M   & 23.59M   & 23.55M   & 23.91M   & 29.24M    \\
Inference Time (s)   & 0.4016   & 2.7291    & 0.4355   & 0.2459   & 0.3041   & 0.1319   & 0.3657   \\
\bottomrule 
\end{tabular}%
\end{table*}
%%%%%%%%%%%%%%%%%%%%%%%%%%%%%%%

%%%%%%%%%%%%%%%%%%%%%%%%%%%%%%%
\begin{figure*}[t]
\vspace{-8pt}
    \centering
    \includegraphics[width=0.99\linewidth]{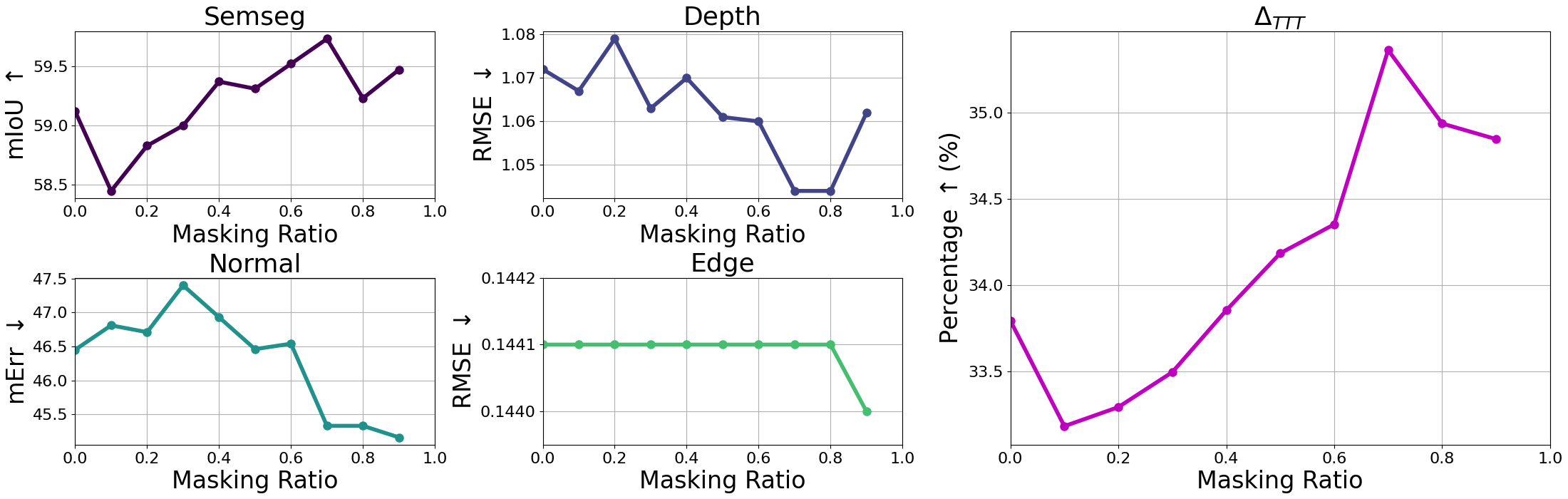}
    \caption{Ablation study on the masking ratio of S4T. We evaluate performance under the domain shift from Taskonomy to NYUD-v2.}
    \label{fig:mask_ratio}
\end{figure*}
%%%%%%%%%%%%%%%%%%%%%%%%%%%%%%%

\clearpage
\onecolumn
\section{Derivations of \cref{proposition1}}
\label{append:derivation}
For simplicity, denote the task-specific latent space as \(\{z_{t,i}\}_{i=1}^{n}\) and its masked version as \(\{\tilde{z}_{t,i}\}_{i=1}^{n}\).

\begin{align}
    d(\theta, p(\{z_{t,i}\}_{i=1}^{n}, &y_{t,j})) - d(\theta, p(\{\tilde{z}_{t,i}\}_{i=1}^{n}, y_{t,j})) \\
     = &\mathbb{E}_{p(\{z_{t,i}\}_{i=1}^{n})} [d[p(y_{t,j}|\{z_{t,i}\}_{i=1}^{n}), p_{\theta}(y_{t,j}|\{z_{t,i}\}_{i=1}^{n})]] \\
    & - \mathbb{E}_{p(\{\tilde{z}_{t,i}\}_{i=1}^{n})} [d[p(y_{t,j}|\{z_{t,i}\}_{i=1}^{n}), p_{\theta}(y_{t,j}|\{\tilde{z}_{t,i}\}_{i=1}^{n})]] \\
     = &\mathbb{E}_{p(\{z_{t,i}\}_{i=1}^{n}, \{\tilde{z}_{t,i}\}_{i=1}^{n})} [d[p(y_{t,j}|\{z_{t,i}\}_{i=1}^{n}), p_{\theta}(y_{t,j}|\{z_{t,i}\}_{i=1}^{n})]] \\
    & - \mathbb{E}_{p(\{z_{t,i}\}_{i=1}^{n}, \{z_{t,i}\}_{i=1}^{n}
\{\tilde{z}_{t,i}\}_{i=1}^{n})} [d[p(y_{t,j}|\{z_{t,i}\}_{i=1}^{n}), p_{\theta}(y_{t,j}|\{\tilde{z}_{t,i}\}_{i=1}^{n})]] \\
    \leq & \mathbb{E}_{p(\{z_{t,i}\}_{i=1}^{n}, \{\tilde{z}_{t,i}\}_{i=1}^{n})} [d[p_{\theta}(y_{t,j}|\{z_{t,i}\}_{i=1}^{n}), p_{\theta}(y_{t,j}|\{\tilde{z}_{t,i}\}_{i=1}^{n})]] \\
    & + \mathbb{E}_{p(\{z_{t,i}\}_{i=1}^{n}, \{\tilde{z}_{t,i}\}_{i=1}^{n})} [d[p(y_{t,j}|\{z_{t,i}\}_{i=1}^{n}), p(y_{t,j}|\{\tilde{z}_{t,i}\}_{i=1}^{n})]]
    \label{eq:tri}
\end{align}
The \cref{eq:tri} follows from the triangle inequality.

\noindent
Rearranging the above equation results in the following inequality.:
\begin{align}
    d(\theta, p(\{z_{t,i}\}_{i=1}^{n}, y_{t,j})) &\leq d(\theta, p(\{\tilde{z}_{t,i}\}_{i=1}^{n}, y_{t,j})) \\
     &+ \mathbb{E}_{p(\{z_{t,i}\}_{i=1}^{n}, \{\tilde{z}_{t,i}\}_{i=1}^{n})} [d[p_{\theta}(y_{t,j}|\{z_{t,i}\}_{i=1}^{n}), p_{\theta}(y_{t,j}|\{\tilde{z}_{t,i}\}_{i=1}^{n})]] \\
     &+ \mathbb{E}_{p(\{z_{t,i}\}_{i=1}^{n}, \{\tilde{z}_{t,i}\}_{i=1}^{n})} [d[p(y_{t,j}|\{z_{t,i}\}_{i=1}^{n}), p(y_{t,j}|\{\tilde{z}_{t,i}\}_{i=1}^{n})]]
     \label{eq:dist}
\end{align}

\noindent
In the multi-task setting, we apply \cref{eq:dist} to each task as follows:
\begin{align}
    \sum_{j=1}^{n} d(\theta, p(\{z_{t,i}\}_{i=1}^{n}, y_{t,j})) \leq& \sum_{j=1}^{n} d(\theta, p(\{\tilde{z}_{t,i}\}_{i=1}^{n}, y_{t,j})) \\
    &+ \sum_{j=1}^{n} \mathbb{E}_{p(\{z_{t,i}, \tilde{z}_{t,i}\}_{i=1}^{n})} [d[p_{\theta}(y_{t,j}|\{z_{t,i}\}_{i=1}^{n}), p_{\theta}(y_{t,j}|\{\tilde{z}_{t,i}\}_{i=1}^{n})]] \\
    &+ \sum_{j=1}^{n} \mathbb{E}_{p(\{z_{t,i}, \tilde{z}_{t,i}\}_{i=1}^{n})} [d[p(y_{t,j}|\{z_{t,i}\}_{i=1}^{n}), p(y_{t,j}|\{\tilde{z}_{t,i}\}_{i=1}^{n})]] \label{eq:dist_target}\\
    \leq& \sum_{j=1}^{n} d(\theta, p(\{\tilde{z}_{t,i}\}_{i=1}^{n}, y_{t,j})) \\
    &+ \sum_{j=1}^{n} \mathbb{E}_{p(\{z_{t,i}, \tilde{z}_{t,i}\}_{i=1}^{n})} [d[p_{\theta}(y_{t,j}|\{z_{t,i}\}_{i=1}^{n}), p_{\theta}(y_{t,j}|\{\tilde{z}_{t,i}\}_{i=1}^{n})]] \\
    &+C\cdot\sum_{j=1}^{n} \mathbb{E}_{p(\{z_{s,i}, \tilde{z}_{s,i}\}_{i=1}^{n})} [d[p(y_{s,j}|\{z_{s,i}\}_{i=1}^{n}), p(y_{s,j}|\{\tilde{z}_{s,i}\}_{i=1}^{n})]] \label{eq:dist_source}
\end{align}
The inequality between \cref{eq:dist_target} and \cref{eq:dist_source} holds under Assumption~\ref{assumption1} with a scaling factor \( C \), which asserts that task relations remain proportionally consistent between tasks and their masked counterparts.
With a properly chosen masking ratio, TBS effectively captures task relations in the source domain, as ensured by Assumption~\ref{assumption2}. Consequently, \cref{eq:dist_source} approaches zero, aligning with the training objective in the source domain. Therefore, the following inequality holds:
\begin{align}
    \sum_{j=1}^{n} d(\theta, p(\{z_{t,i}\}_{i=1}^{n}, y_{t,j})) \leq& \sum_{j=1}^{n} d(\theta, p(\{\tilde{z}_{t,i}\}_{i=1}^{n}, y_{t,j})) \\
    &+ \sum_{j=1}^{n} \mathbb{E}_{p(\{z_{t,i}, \tilde{z}_{t,i}\}_{i=1}^{n})} [d[p_{\theta}(y_{t,j}|\{z_{t,i}\}_{i=1}^{n}), p_{\theta}(y_{t,j}|\{\tilde{z}_{t,i}\}_{i=1}^{n})]]
\end{align}